\documentclass{article}
\usepackage{ijcai22}

\usepackage{tabularx} 
\usepackage{times}
\usepackage{soul}
\usepackage{url}
\usepackage[hidelinks]{hyperref}
\usepackage[utf8]{inputenc}
\usepackage[small]{caption}
\usepackage{graphicx}
\usepackage{amsmath}
\usepackage{amsthm}
\usepackage{booktabs}
\usepackage{algorithm}
\usepackage{algorithmic}
\usepackage{fancyhdr}
\usepackage[table,dvipsnames]{xcolor}
\newcommand{\cmt}[1]{\ignorespaces}
\newcommand{\ncmt}[1]{\ignorespaces}
\newcommand{\cmtn}[1]{#1}

\title{Learning Label Initialization for \emph{Time-Dependent} Harmonic Extension}

\author{
    Amitoz Azad
    \affiliations
    University of Caen Normandy
    \emails
    amitoz.sudo@gmail.com
}

\begin{document}
\maketitle

\begin{abstract}
\cmt{In this paper, we consider a \emph{time-dependent} version of the Dirichlet problem on graphs and show how to improve its solution by learning proper initialization vector on the unlabeled nodes.}
Node classification on graphs can be formulated as the Dirichlet problem on graphs where the signal is given at the labeled nodes, and the harmonic extension is done on the unlabeled nodes. 
This paper considers a \emph{time-dependent} version of the Dirichlet problem on graphs and shows how to improve its solution by learning the proper initialization vector on the unlabeled nodes.
Further, we show that the improved solution is at par with state-of-the-art methods used for node classification.
Finally, we conclude this paper by discussing the importance of parameter \emph{t}, pros, and future directions.

\end{abstract}

\section{Introduction}
\label{Introduction}

Node classification on graphs has various real-world applications~\cite{song2021graph} ranging from computer vision, natural language processing, social network, biomedical sciences. Often it is done in a semi-supervised setting in which the labeled data is significantly less than the unlabeled data, and the goal is to use both the labeled and unlabeled data to obtain a good classification score. This makes it an ideal approach for the problems where labeled data is not cheap or scarcely available.

Many state-of-the-art methods for node classification use machine learning on graphs~\cite{kipf2016semi,velivckovic2017graph,monti2017geometric}. Graph Neural Networks (GNNs) are emerging as powerful tools for graph representation learning. The earliest method for node classification on graphs used Laplacian regularization~\cite{zhu2003semi,belkin2004regularization,zhou2005regularization}. The Laplacian regularization can also be formulated as a PDE approach on graphs (Dirichlet problem~\eqref{eq:pde}). Several non-trivial Laplacians have been proposed, and the Dirichlet problem associated with each one of them could be used for node classification~\cite{calder2018game,kyng2015algorithms,calder2020properly,el2016asymptotic}. In this work, we consider the Dirichlet problem associated with normalized Laplacian as defined in~\cite{zhou2005regularization}. More specifically, we consider the time-dependent formulation of the Dirichlet problem~\eqref{eq:time-pde}  and argue about making its solution as competitive as state-of-the-art methods for transductive learning (Table~\ref{tb:2}).

The triggering point of this work was the comparison of GCN's~\cite{kipf2016semi} accuracy and the solution of the Dirichlet problem~\eqref{eq:time-pde} on handcrafted \emph{knn} graphs  for MNIST~\cite{lecun1998gradient},~FMNIST~\cite{xiao2017fashion}, and the popular citation graphs Cora, Citeseer, Pubmed~\cite{sen2008collective}.
We observed that while the solutions of the PDE and the GCN are comparable for node classification on the handcrafted \emph{knn} graphs, the latter performed significantly better on the citation graphs (Table~\ref{tb:1}).
For the handcrafted graphs, the nodes correspond to the images, and node features are the pixel values of the images.
The plausible reason for the similar performances on handcrafted \emph{knn} graphs would be that edges connect the similar nodes\cmt{represent a good measure of similarity between the nodes}, as the euclidean distance between the images happens to be a good metric for the graph construction in this case.
The citation graphs are given to us by default; the nodes represent documents, edges correspond to citation links between the documents rather than their similarity, and each node has a high dimensional bag-of-words feature vector.
The apparent reason for the superior performance of the GCN on citation graphs would be that not only does it take the graph as an input, but it also takes the node features as input and utilizes their hidden representations.

\paragraph{Main contributions.} Recent work has shown that GNNs are good at utilizing the node feature information~\cite{faber2021should}. Since the GCN beats the PDE-based approach on \mbox{citation} graphs by employing the node features along with the graph. This led us to ask: \emph{How to make the solution of the PDE~\eqref{eq:time-pde} \cmt{time dependent Dirichlet problem} at par with the GCN by utilizing the node feature information?}
\begin{itemize}
\item We show that this is achieved by utilizing the node features to learn the initial condition~\eqref{eq:ic} of the PDE (Figure~\ref{fg:1}).
\item We benchmark the improved solution against the several state-of-the-art methods for node classification. Our approach yields competitive results (Table~\ref{tb:2}, Row 08).
\item Although we learn the initialization vector to boost the performance of the PDE, further marginal gains are also added by learning the graph weights (Table~\ref{tb:2}, Row 09).
\item Unlike GNNs, which need to be retrained every time to incorporate new labels, this approach can easily incorporate new labels after training. We demonstrate this by incorporating validation labels (Table~\ref{tb:2}, Row 11). \mbox{Importantly}, we do not use the new labels to retrain. 
\end{itemize}
\section{Background}
\cmt{PdE (\textbf{partial \emph{difference} equations}) on graphs~\cite{neuberger2006nonlinear,gilboa2009nonlocal} and GSP~\cite{ortega2018graph} (\textbf{graph signal processing}) are two emerging fields of handling regular and irregular type of data with lots of applications.   Non-local signal processing is an emerging}
In this section, we review notation on graphs and formulate node-classification as the \emph{time-dependent} Dirichlet problem on graphs.  

\subsection{Preliminaries}
A weighted graph $ G = (V, E, w) $ consists of a finite set $ V $ of $ N $ nodes 
and  a finite set $ E $ $ \subset V \times V $ of edges. Let $ (u, v) $ be an edge connecting the nodes $ u $ and $ v $. A weighted graph is associated with a weight function $ w: V \times V \rightarrow [0,1] $. It  represents a measure of similarity between two nodes. The set of edges according to the weight function is given as: $ E = \{(u, v) | w (u, v) \neq 0 \} $. The set of nodes in the neighborhood of  node $ u $ is denoted as $ N (u) $. The notation $ v \in N (u) $ means node $ v $ is in the neighborhood of node $ u $, \emph{i.e.}  $ N (u) = \{v \in  V | (u, v) \in E \} $. In this paper, we consider symmetric graphs \emph{i.e.} $ w (u, v) = w (v, u) $ and $ (u, v) \in E \leftrightarrow (v, u) \in E $. The degree of a node $ u $ is given as: $ \delta_{w}(u) = \sum_{v \in N (u)} w (u, v) $. 

Let  $ H (V) $ be a Hilbert space of the real-valued function on the nodes of the graph. A function  $ f: V \rightarrow R $ of $ H (V) $, represents a signal on a node and assigns a real value $ f (u) $ to each node $ u \in V $. Similarly, let $ H (E) $ be a Hilbert space of the real-valued function defined on the edges. These two spaces are equipped with the following inner products:
\begin{equation}
\label{eq:inner}
\begin{split}
{\langle f,g \rangle}_{H(V)}&=\sum_{u\in V} f(u) g(u),  \quad \forall f, g \in H(V)\\ 
{\langle F, G \rangle}_{H(E)}&=\sum_{(u,v)\in E}\!\!F(u,v)G(u,v), \ \forall F,G \in H(E)
\end{split}
\end{equation}

The weighted graph gradient (or difference) operator  $\nabla_{w} : H(V) \rightarrow H(E)$ is defined as:
\begin{align}
\label{eq:grad}
  \nabla_{w}(f)(u,v) := \sqrt{\frac{w(u,v)}{\delta_{w}(v)}}f(v) - \sqrt{\frac{w(u,v)}{\delta_{w}(u)}}f(u)
\end{align}
The norm of the graph gradient on each node is given as:
\begin{align}
   \|\nabla_{w} (f)(u)\| = \left(\sum_{v\in N(u)} (\nabla_{w} (f))^{2}(u,v)\right)^{\tfrac{1}{2}}
\end{align}
The \emph{Dirichlet} energy associated with the signal $f(u)$ is defined as :
\begin{equation}
   J(f) := \frac{1}{2} \sum_{v \in V}\|\nabla_{w} (f)(u)\|^{2}
\end{equation}
Often it is also known as \emph{Tikhonov regularization}~\cite{belkin2004regularization}.
The weighted graph divergence operator $div_{w} : H(E) \rightarrow H(V)$ satisfies the discrete version of stokes law:
\begin{multline}
\label{eq:stoke}
  \langle \nabla_{w}(f), F \rangle_{H(E)}= \langle f, -div_{w} (F)\rangle_{H(V)}\\
\text{where} \ f \in H(V) \ \text{and} \   F \in H(E)
\end{multline}
Using~\eqref{eq:inner},~\eqref{eq:stoke} and let $f=\mathbf{1}_{\{u\}}$, it can be shown that divergence operator is given as:
  \begin{align}
  \label{eq:div}
     div_{w}(F)(u) = \sum_{v\in N(u)}\sqrt{\frac{w(u,v)}{\delta_{w}(u)}}(F(u,v) - F(v,u))
  \end{align}
The graph Laplacian is defined as $\Delta_{w}: H(V) \rightarrow H(V)$:
\begin{align}
\Delta_{w} (f)(u) := \frac{1}{2}div_{w}(\nabla_{w} (f))(u)
\end{align}
Using the definitions of $\nabla_{w}$ and $div_{w}$, the Laplacian is given as: 
\begin{align}
  \Delta_{w}(f)(u) = \sum_{v\in N(u)} w(u,v) \left(\frac{f(v)}{\sqrt{\delta_{w}(u)\delta_{w}(v)}} - \frac{f(u)}{\delta_{w}(u)}\right)
\end{align}

\subsection{Node Classification}
Let us consider the problem of node classification on graphs in transductive settings for a multi-class problem. \cmt{As a side note, it could also be viewed as a problem of signal interpolation on graphs.}
Let $V_{0}=\{u_{1}, u_{2}, ..., u_{m}\}$ be a subset of nodes in $V$ over which the labels are given as $y_{1},y_{2}, ..., y_{m} \in \mathbf{\{e_{1}, e_{2}, ..., e_{k}\}}$, where $\mathbf{e_{i}} \in R^{k}$  represents the $i^{th}$ class out of $k$ classes (one-hot vector).
The goal of transductive learning is to extend these labels to the unlabeled nodes.
One way to achieve this is via solving the following inverse problem:
\begin{equation}
\label{eq:var-problem}
   \underset{f(u)}{\arg \min} \left\{\sum_{u\in V} \tfrac{1}{2} \|\nabla_w (f)(u)\|^{2} \  \text{:} \ f(u) = g(u) \quad \forall u \in V_{0}\right \} 
\end{equation}
Here, $g: V_{0} \rightarrow R^{k}$ is the label function on the nodes, such that $g(u) \in \mathbf{\{e_1,e_2, ..., e_k\}}$.
The other way to extend the known labels is via solving the following PDE (\emph{a.k.a the Dirichlet problem}) on the graph, which is obtained through the E-L equation of the above optimization:
\begin{equation}
\label{eq:pde}
\begin{split}
    \Delta_{w} (f)(u) &= 0  \quad \forall u \in V \setminus V_0 \\
    f(u) &= g(u)  \quad \forall u \in V_{0}
\end{split}
\end{equation}
It is worth remembering that the above equation is vector-valued PDE, with $k$ components.
The final label decision for the unlabeled node $u$ is determined by the largest component of $f(u)$:
\begin{align}
l(u) = \underset{j\in\{1,...,k\}}{\arg\max}\{f_{j}(u)\}
\end{align}

\subsection{Connection With Label Propagation}
We now digress briefly to mention the connection with the widely-used label propagation algorithm~\cite{zhu2005semi}. For a symmetric graph and under somewhat different definitions of gradient and divergence 
\begin{equation}
\begin{split}
\nabla_{w}(f)(u,v) &= \sqrt{w(u,v)}(f(v) - f(u))\\
div_{w}(F)(u) &= \sum_{v\in V}\sqrt{w(u,v)}(F(u,v) - F(v,u)),
\end{split} 
\end{equation}
the Laplacian on the left-hand side of~\eqref{eq:pde} yields the classic unnormalized (\emph{a.k.a combinatorial}) Laplacian ~\cite{chung1997spectral}. The Dirichlet problem then becomes:
\begin{align}
   \sum_{v\in N(u)} &w(u,v) (f(v) - f(u)) = 0, \quad \forall u \in V \setminus V_0 \nonumber \\
    &f(u) = g(u), \quad \forall u \in V_{0}
\end{align}
The above equation can be solved by the Jacobi's iterative method. Which yields the classic label propagation:
\begin{align}
    f^{n+1}(u) &= \frac{\sum_{v \in N(u)}w(u,v) f^{n}(v)}{\sum_{v\in N(u)}w(u,v)}, \quad \forall u \in V \setminus V_{0} \nonumber \\
    f^{n+1}(u) &= g(u),  \quad \forall u \in V_{0}
\end{align}

\subsection{Time-Dependent Dirichlet Problem}
A traditional \cmt{An effortless} approach to solve E-L equation of a variational problem is to make it \emph{time-dependent}. 
A celebrated example of this is \emph{R.O.F.} PDE, which is used in image processing for anisotropic denoising~\cite{rudin1992nonlinear}.
\cmt{Classic example of this technique of converting E-L to a time dependent PDE is R.O.F PDE used in image processing for anisotropic denoising~\cite{rudin1992nonlinear}.}
\cmt{and few other ().}
Formulating the Dirichlet problem in~\eqref{eq:pde} as time-dependent, it becomes: 
\begin{subequations}
\label{eq:time-pde}
\begin{align}
\frac{\partial f(u,t)}{\partial t}  &= \Delta_{w}(f)(u,t) &\forall u \in V \setminus V_{0}\label{eq:main}\\
f(u,t) &= g(u) &\forall u \in V_0\label{eq:bc}\\
f(u,0) &=  \psi_{0}(u) &\forall u \in V\label{eq:ic}
\end{align}
\end{subequations}
At steady-state, when $\frac{\partial f}{\partial t}\rightarrow 0$, the solution of above equation is equivalent to that of~\eqref{eq:pde}.
Basically, it is a gradient flow in the direction of minimizing the Dirichlet energy while satisfying the boundary constraint. 
The above equation can also be viewed as heat diffusion with non-homogeneous boundary condition on graphs.
Notice that the Dirichlet problem in~\eqref{eq:pde} had only a boundary condition, but now it also has an initial condition, denoted as $\psi_0$.

A common choice of setting $\psi_0$ is to let the signal be $g(u)\in R^{k}$ (one-hot vectors) on the labeled nodes and zero vectors elsewhere.
It can be seen as an initial unnormalized probability vector on nodes. 
In this paper, we choose to address it as `front $\psi_0$' since the approach in~\eqref{eq:time-pde} is similar to level-set (front-propagation) methods in computer vision~\cite{osher1993level,osher2003geometric}.\footnote{Certain Dirichlet problems can be formulated as \mbox{\emph{time-dependent}} PDE with no boundary condition.}
\cmt{The proposed approach of this paper to learn $\psi_0$ could be seen as learning front initialization of leve-set evolution on graph.}
\cmt{Mention that maximum value principle is satisfied for all t.}

\section{Experiments and Results}
 This section walks through the research questions (RQs) and discusses the experiments done to answer them. \ncmt{The reader is advised to follow the Technical Appendix side by side\cmt{}.}
 
\begin{table}[t!]
\centering
\begin{tabular}{lll}
\toprule
Dataset  & \textbf{Eq15} & \textbf{GCN}\\
\midrule
\textbf{MNIST}   & 93.2 \cmtn{$\pm$ 0.0}    & 91.3 \cmtn{$\pm$ 0.2}     \\
\textbf{FMNIST}  & 76.0 \cmtn{$\pm$ 0.0}  & 77.4 \cmtn{$\pm$ 0.2}     \\
\rowcolor{Salmon}
\textbf{Cora}    & 72.5 \cmtn{$\pm$ 0.0} & {81.5 \cmtn{$\pm$ ---}}\textsuperscript{\textdagger}     \\
\rowcolor{Salmon}
\textbf{Citeseer} & 49.7 \cmtn{$\pm$ 0.0} & {70.3  \cmtn{$\pm$ ---}}\textsuperscript{\textdagger}    \\
\rowcolor{Salmon}
\textbf{Pubmed}   & 72.5 \cmtn{$\pm$ 0.0} & {79.0  \cmtn{$\pm$ ---}}\textsuperscript{\textdagger}   \\
\bottomrule
\end{tabular}
\caption{
Performances of~\eqref{eq:time-pde} and the GCN on handcrafted and citation graphs. Observe the significant differences in scores over the latter.\textdagger Values taken from the GCN paper.
}
\label{tb:1}
\end{table}

 \paragraph{Code.} The scripts used for the experiments and the instructions to run are available at Github: \url{https://github.com/aGIToz/Learning-Label-Initialization}.
 
\begin{figure*}[t!]
\centering
\includegraphics[width=\textwidth]{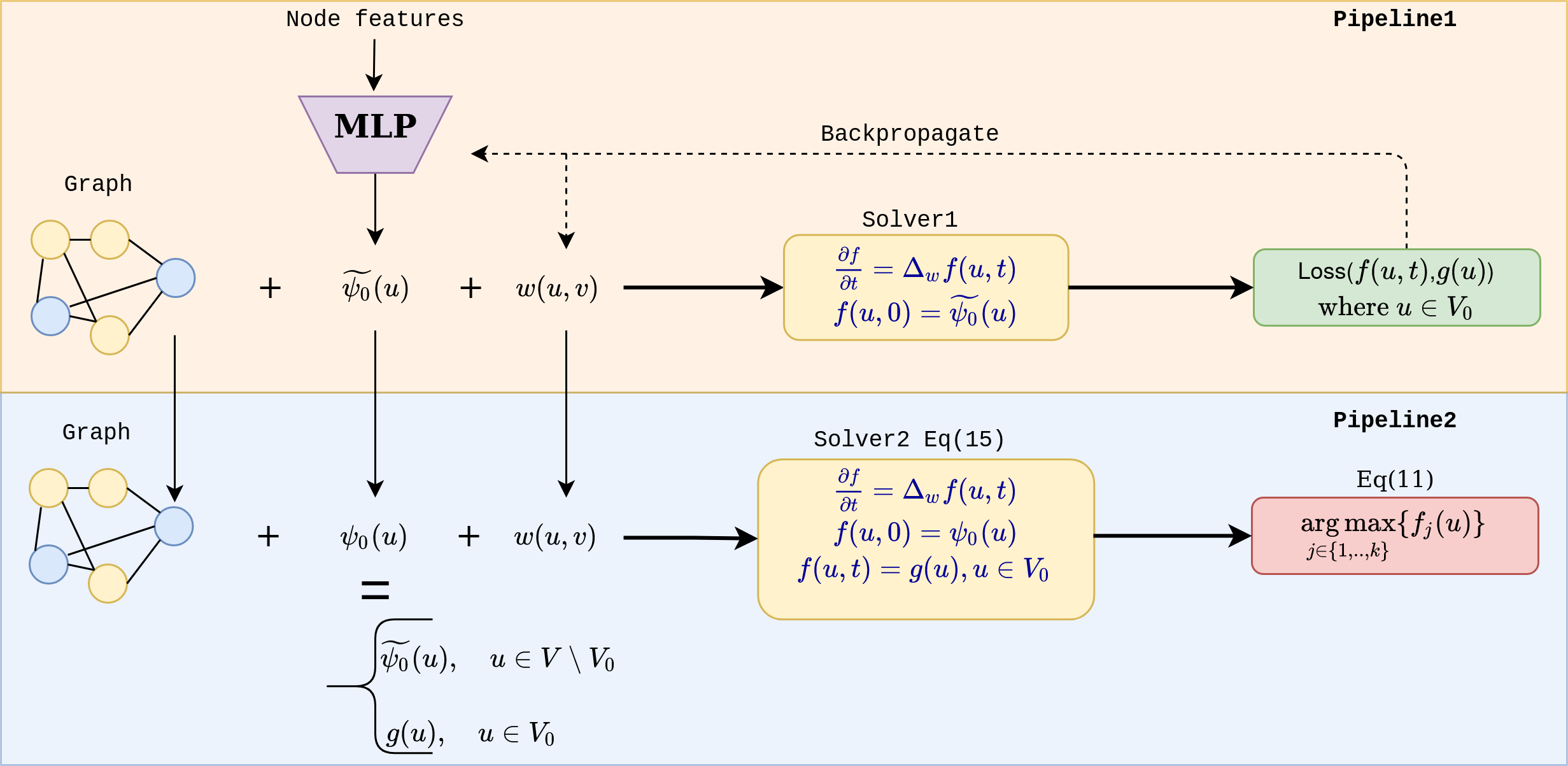}
\caption{Proposed architecture to learn the front $\psi_0$ on the unlabeled nodes.}
\label{fg:1}
\end{figure*}

\paragraph{RQ1.} \emph{How does the solution of~\eqref{eq:time-pde} compare with the GCN~\cite{kipf2016semi} for semi-supervised node classification?}

\noindent To answer this question, we tested them on handcrafted \emph{(knn)} graphs  and non-handcrafted citation graphs. 

\paragraph{Handcrafted graphs.} We created \emph{knn} graphs for two datasets:
MNIST~\cite{lecun1998gradient} and  F(fashion)-MNIST~\cite{xiao2017fashion}. 
MNIST is a popular dataset of handwritten digits and 
FMNIST is a dataset from Zalando's clothing images; both consist of 70k examples of sizes 28 by 28 and 10 classes. For the graph construction $k$ was set to 10 and we used RBF kernel $w(u,v)=e^{-{d(u,v)^{2}}/{\sigma^2}}$ for weights. \ncmt{More details are shared in the Appendix B.}

\paragraph{Non-handcrafted graphs.} We used the popular citation graphs, which have been widely used for benchmarking Graph nets. These are Cora, Citeseer, Pubmed~\cite{sen2008collective}.
Here the nodes represent documents; edges correspond to citation links between them, and each node has a sparse bag-of-words feature vector and a class label. 
\ncmt{More details about these datasets are shared in Appendix B.}

\paragraph{Software.} We used PyTorch framework~\cite{paszke2019pytorch}.
To run the geometric PDE~\eqref{eq:time-pde}, we used torch-geometric~\cite{fey2019fast} and torchdiffeq~\cite{chen2018neural}. 
Torchdiffeq is a popular ODE solver written in PyTorch. It allows to backpropagate via adjoint sensitivity method~\cite{boltyanskiy1962mathematical} and comes with different numerical schemes. For all the experiments, we used the Dormand-Prince (dopri5) method. It is also recommended by the authors of the library, and it is a common choice in \mbox{NeuralODE} literature (Sec~\ref{sec:rw}).

\paragraph{Split setting.} 
For the MNIST and FMNIST datasets, we kept the seed size (labeled nodes) 20 nodes per class, for validation 500 nodes per class, and the rest for the test set.
The citation graphs come with a prebuilt split (\emph{a.k.a. \mbox{planetoid~split}}), which is very often used for benchmarking node classification with GNNs. 
These datasets come with default training, validation, and test sets. The training sets of the citation graphs were used as seeds (labeled nodes). 
The validation sets of all the datasets were used to tune the hyperparameters of the GCN and the ODE solver.
It was also used to tune $\sigma$ parameter in the RBF kernel in the weights of handcrafted \emph{knn} graphs. 
\ncmt{The precise values can be found in the Appendix B, C.}
The initialization of the front~$\psi_0$ was done as a one-hot vector on the labeled nodes and zero vector on the unlabelled nodes. 

\paragraph{Evaluation.} We report the accuracy on the test nodes for 100 random iterations (Table~\ref{tb:1}). The GCN's results on citation graphs are taken from their paper~\cite{kipf2016semi}.
 For every dataset, the same graph structure was given to the ODE solver and the GCN. {Note that the performance of GCN is sensitive to weight initialization (random iterations) and hence comes with variations, but the solution of~\eqref{eq:time-pde} is deterministic and does not fluctuate for different iterations~(\mbox{Table~\ref{tb:1}}).

\paragraph{Observations.} 
We observe a notable discrepancy between the performances of the GCN and~\eqref{eq:time-pde} for citation graphs compared to the handcrafted \emph{knn} graphs (Table~\ref{tb:1}).
The GCN performs significantly better on the citation graphs.
The likely reason for the poor performance of the PDE on the citation graphs could be attributed to the fact that the edges do not explicitly \cmt{necessarily} translate to the similarity between the nodes on these graphs. In fact, they correspond to citation links between the documents. 
This breaks the premise of LP (label propagation) like algorithms,  which assume that edges and weights correspond to the similarity between the nodes. Often LP like algorithms are evaluated on the graphs constructed using node features, but in this case, the graphs correspond to citation \emph{networks}, and they are already made for us.
The GCN, on the other hand, is able to perform well on citation networks as it uses the graph structure and node features, whereas~\eqref{eq:time-pde} only takes the graph structure as an input. This leads us to our next research question.

\paragraph{RQ2.} \emph{Knowing that the GCN beats the solution of~\eqref{eq:time-pde} by utilizing the node features and the graph structure, could we use the node features and make the solution of~\eqref{eq:time-pde} at par with the GCN on the citation graphs?}

\noindent We show that this is indeed possible by learning the front $\psi_0$~\eqref{eq:ic} on the unlabeled nodes.
Not only does this improve the solution, but the results are made as competitive as several state-of-the-art methods (Table 2).

{
\begin{table*}[t!]
\centering
\begin{tabular*}{\textwidth}{@{\extracolsep{\fill}} lllllll}
\toprule
\textbf{Model} &  \textbf{Cora} & \textbf{Citeseer} & \textbf{Pubmed} & \textbf{Photo} & \textbf{CountyFB} & \textbf{FMAsia} \\
\midrule
01 \textbf{GCN} & {81.5  \cmtn{$\pm\ $ ---}}\textsuperscript{\textdagger} & {70.3 \cmtn{$\pm\ $ ---}}\textsuperscript{\textdagger} & {79.0 \cmtn{$\pm\ $ ---}}\textsuperscript{\textdagger} & 93.4 \cmtn{$\pm$ 0.6} & 76.4 \cmtn{$\pm$ 1.4} & 85.9 \cmtn{$\pm$ 0.7}\\
02 \textbf{GAT} & {\textcolor{blue}{{83.0} \cmtn{$\pm$ 0.7}}}\textsuperscript{\textdagger} & {\textcolor{red}{{72.5} \cmtn{$\pm$ 0.7}}}\textsuperscript{\textdagger} & {79.0 \cmtn{$\pm$ 0.3}}\textsuperscript{\textdagger} & \textcolor{black}{94.1 \cmtn{$\pm$ 0.7}} & 76.3 \cmtn{$\pm$ 1.6}& \textcolor{OliveGreen}{87.0 \cmtn{$\pm$ 0.8}} \\
03 \textbf{MoNet} & {81.7 \cmtn{$\pm$ 0.5}}\textsuperscript{\textdagger} & 70.3 \cmtn{$\pm$ 0.9} & {78.8 \cmtn{$\pm$ 0.4}}\textsuperscript{\textdagger} & \textcolor{blue}{94.3 \cmtn{$\pm$ 0.3}} & 79.4 \cmtn{$\pm$ 1.5} & 86.1  \cmtn{$\pm$ 1.0}\\
04 \textbf{C\&S} \textbf{(MLP)} &  80.0 \cmtn{$\pm$ 0.9} & 67.3 \cmtn{$\pm$ 1.3} & 75.9 \cmtn{$\pm$ 1.9} & \textcolor{OliveGreen}{94.2 \cmtn{$\pm$ 0.5}} & \textcolor{OliveGreen}{82.2 \cmtn{$\pm$ 1.2}} & \textcolor{red}{87.3 \cmtn{$\pm$ 0.7}}\\
05 \textbf{CGNN(w/o)} &  {82.7 \cmtn{$\pm$ 0.7}}\textsuperscript{\textdagger} & {71.5 \cmtn{$\pm$ 1.1}}\textsuperscript{\textdagger} & {\textcolor{blue}{{81.7}  \cmtn{$\pm$ 0.5}}}\textsuperscript{\textdagger} & 93.3 \cmtn{$\pm$ 0.2} & 80.5 \cmtn{$\pm$ 1.4} & 85.3 \cmtn{$\pm$ 0.4} \\
06 \textbf{CGNN(w/)} &  {82.8 \cmtn{$\pm$ 0.6}}\textsuperscript{\textdagger} & {{72.1} \cmtn{$\pm$ 0.8}}\textsuperscript{\textdagger} & {\textcolor{blue}{{81.7} \cmtn{$\pm$ 0.8}}}\textsuperscript{\textdagger} & 92.2 \cmtn{$\pm$ 0.7} & 81.9 \cmtn{$\pm$ 0.9} & 85.4  \cmtn{$\pm$ 0.3}\\
07 \textbf{Eq15} & {72.5}  \cmtn{$\pm$ 0.0}  & 49.7 \cmtn{$\pm$ 0.0}& 72.5 \cmtn{$\pm$ 0.0}& 91.3 \cmtn{$\pm$ 0.5}& 81.4 \cmtn{$\pm$ 1.8}& 86.7 \cmtn{$\pm$ 0.8}\\
08  \textbf{Eq15} {\small learned $\psi_0$ }  & \textcolor{OliveGreen}{{82.9} \cmtn{$\pm$ 0.7}} \cmt{\small 3s} & \textcolor{OliveGreen}{72.2 \cmtn{$\pm$ 0.6}} \cmt{\small 7s} & \textcolor{OliveGreen}{{81.1} \cmtn{$\pm$ 0.9}} \cmt{\small 31s} & \textcolor{blue}{{94.3} \cmtn{$\pm$ 0.5}} \cmt{\small 4.3m} & \textcolor{blue}{82.4 \cmtn{$\pm$ 1.5}} \cmt{\small 14s} & \textcolor{OliveGreen}{87.0   \cmtn{$\pm$ 0.9}} \cmt{\small 30s}\\
09 \textbf{Eq15} {\small learned $\psi_0$ \& ${w}$} & \textcolor{red}{{83.5} \cmtn{$\pm$ 0.7}} & \textcolor{blue}{{72.4} \cmtn{$\pm$ 0.7}} & \textcolor{red}{{81.9} \cmtn{$\pm$ 0.9}} & \textcolor{red}{{94.4} \cmtn{$\pm$ 0.5}} & \textcolor{red}{82.6 \cmtn{$\pm$ 1.8}} & \textcolor{blue}{87.1 \cmtn{$\pm$ 0.8}} \\
\midrule
10 \textbf{Eq15} {\small inc. val} & {79.5} \cmtn{$\pm$ 0.0} & 60.4 \cmtn{$\pm$ 0.0} & {78.4} \cmtn{$\pm$ 0.0} & {92.4}  \cmtn{$\pm$ 0.6} & 82.8 \cmtn{$\pm$ 1.4} & 87.2   \cmtn{$\pm$ 0.7}\\
11 \textbf{Eq15} {\small inc. val, ($\psi_0$,$w$)} & {87.2} \cmtn{$\pm$ 0.0} & 73.3 \cmtn{$\pm$ 0.0} & {83.5} \cmtn{$\pm$ 0.0} & {95.5}  \cmtn{$\pm$ 0.5} & 85.3 \cmtn{$\pm$ 1.4} & 88.6   \cmtn{$\pm$ 0.8}\\
\bottomrule                             
\end{tabular*}
\caption{Test accuracy over different datasets.\textdagger Values taken from the original papers. Coloring scheme: red-$1^{st}$, blue-$2^{nd}$, green-$3^{rd}$ .}
\label{tb:2}
\end{table*}
}

\paragraph{Proposed model.} Figure~\ref{fg:1} shows the proposed architecture to enhance the solution of~\eqref{eq:time-pde}. 
It is composed of two Pipelines. 
Pipeline1 is a training pipeline; it is used for learning the $\psi_0$ on unlabeled nodes.
Node features are plugged into an MLP to generate an initial estimate of the front ($\widetilde{\psi_0}$), which is further given as an input to Solver1.
Apart from $\widetilde{\psi_0}$, the Solver1 also takes the graph structure and graph weights $w(u,v)$ as inputs.
The graph weights are also made learnable.
Although not shown in the Figure~\ref{fg:1}, the graph weights are learned in such a way to ensure they are symmetric $w(u,v)=w(v,u)$, and between zero and one.
The loss function in Pipeline1 could be seen as a way to enforce the boundary condition~\eqref{eq:bc};
one could argue that such reasoning is motivated by the inspiring work in PINNs~\cite{raissi2019physics}, where a boundary condition of a PDE is turned into a loss function. 
Pipeline2 corresponds to the evaluation pipeline. 
After every epoch, the values of $\widetilde{{\psi_0}}$ and $w(u,v)$ are updated via full-batch gradient descent, and \cmt{; they are then }fed into Solver2 to evaluate the PDE on the validation set and for saving the front and the weights.
Pipeline2 could be regarded as the final model as we use it in the end, to evaluate the performance on the test set, and it can be separately deployed once the front and the weights are learned.
Note the relationship between $\psi_0$ and $\widetilde{\psi_0}$. For the unlabeled nodes, $\psi_0$ vector is equal to $\widetilde{\psi_0}$ vector, and for the labeled nodes, $\psi_0$ vector is equal to the $g(u)$ vector (one-hot). 
Unlike Solver1, the Solver2 keeps the signal on the labeled nodes as $g(u)$ throughout the evolution of~\eqref{eq:time-pde} from 0 to $t$.\footnote{
The reader may be intrigued about the need for Solver2. Why not put the boundary condition~\eqref{eq:bc} in Solver1 and minimize the loss function? It is not possible as the loss will remain zero.}

\paragraph{Baselines.} We compare the improved performances of~\eqref{eq:time-pde} \cmt{with learned $\psi_0$} with three graph networks: Graph Convolutional Network (GCN)~\cite{kipf2016semi}, Graph Attention Network (GAT)~\cite{velivckovic2017graph}, Mixture Model Networks (MoNet)~\cite{monti2017geometric}. 
Additionally, we compare to a hybrid model Correct and Smooth (CS)~\cite{huang2020combining}
and to Continuous Graph Neural Networks \mbox{(CGNN)~\cite{xhonneux2020continuous}} with and without feature mixing.
The latter two models are discussed briefly in Sec~\ref{sec:rw}.

\paragraph{Datasets.} 
Apart from the citation graphs, we include three more real-world datasets: Amazon Photo~\cite{shchur2018pitfalls}, CountyFB~\cite{jia2021unifying}, FMAsia~\cite{rozemberczki2020characteristic}. In Amazon Photo nodes represent items, and edges represent if two items are frequently bought together. 
Node feature is product review as a bag-of-words vector, and the task is to map the nodes to their respective product category.
In CountyFB dataset, the nodes are US Counties, and edges come from FacebookSocialConnectedness index. Nodes have demographic features (income, migration, education ...) and few features from Facebook friendships. The task is to predict the 2016 election outcome (Republican vs Democrat).
The nodes in FMAsia dataset represents users of streaming service LastFM in Asia, and edges are friendships between them. The task is to predict the home country of the users.

\paragraph{Evaluation setting.} We use the standard planetoid split setting for the citation graphs and report the performance over 100 random iterations.  The planetoid split amounts to just 5\% of seeds (training set) per class. 
Small training splits in graph datasets could produce results sensitive to 
the choice of the split~\cite{shchur2018pitfalls}.
Therefore, for Photo, CountyFB, and FMAsia, we keep the split size 40\%~(train), 40\%~(valid), 20\%~(test) per class and evaluate the performance for 10~different splits. \footnote{Each split is generated by torch.random.seed from 0 to 9.}  

\paragraph{Training setting.} We used Adam or RMSprop optimizer in all our experiments. 
The optimization was done with full-batch gradient descent.
The L2 weight regularization and dropouts were also used in all optimizations.
Most of the MLP used in the architecture (Figure~\ref{fg:1}) are with one or two hidden channels with ReLU activations.
The loss function was cross entropy.
The hyperparameters of the Solver1 and Solver2 were kept same.
The Dormand-Prince (dopri5) numerical scheme was used for all the experiments related to~\eqref{eq:time-pde}.
The  validation splits were used for tuning the hyperparams and for the final model selection (\emph{saving the front and the weights}).
Hyperparameter optimization was done using a large random search (coarse to fine). 
We used either categorical distribution or uniform distribution on the hyperparams while running the random search. 
The search was done using NVIDIA 1080Ti and P100 GPUs.\footnote{The final evaluation was done on 1080Ti.} \ncmt{The precise values of all the hyperparams are shared in the \mbox{Appendix~D} \mbox{(including the other models)}.}

\paragraph{Observations.}  
We see that the proposed approach significantly improves the performance of~\eqref{eq:time-pde} and makes it competitive with several other methods (Table~\ref{tb:2}).
Row~07 corresponds to the classical solution of~\eqref{eq:time-pde} (with zeros and one-hot vectors initialization).
Row~08 shows the improvement in the solution with the learned front $\psi_0$. 
And Row~09 shows further slight gains by learning the weights along with the front~$\psi_0$.
\cmt{Notice the variations in Row 07 onwards, even though (15) is inherently deterministic.}
Notice that while~\eqref{eq:time-pde} is inherently deterministic, there are variations from Row 07 onwards. 
These variations can be attributed to the weight initializations in MLP (Pipleine1), and the fact that evaluation is done for the multiple test splits for Photo, CountyFB, and FMAsia datasets.

\paragraph{RQ3.} \emph{How to incorporate new labels in the final model once the training has been ended?}

\noindent This can be quickly done by including the new labels in the boundary condition~\eqref{eq:bc} and then updating a saved front $\psi_0$\footnote{Learned by training on the training set only.} by assigning one-hot vectors to the new labels and \mbox{finally} running Pipeline2.
Row 10 (Table~\ref{tb:2}) corresponds to the solution of~\eqref{eq:time-pde}, including the validation labels in~\eqref{eq:bc} and with the default initialization of the front. Row~11 corresponds to the solution of~\eqref{eq:time-pde} by updating a saved front~$\psi_0$ and using saved weights.
\cmt{The row11 corresponds to solution of the PDE by updating the learned front $\psi_0$. 
Note that the $\psi_0$ is learned only using the training set. 
Learning $\psi_0$ with validation label would require retraining of the model.}
\section{Discussion}
In this section, we discuss some relevant works and conclude this work with key take-homes.

\subsection{Related Work}
\label{sec:rw}
\paragraph{Hybrid models.} The Correct \& Smooth (C\&S)~\cite{huang2020combining} paper is relevant to our work. 
Their work focuses on first generating base predictions by training an MLP, which is then followed by a correction step (error propagation) and subsequently a label smoothing step.
All three steps are done separately.
The correction and smoothing steps are done using the iterative scheme in~\cite{zhou2004learning}.
On the other hand, we learn the base predictions (the front $\psi_0$), but there is no error propagation step, and the final smoothing is done via the PDE~\eqref{eq:time-pde} in Pipeline2.
Also, unlike their work, learning the base prediction is followed by a smoothing step in Pipeline1 (Solver1) and the MLP is learned in an end-to-end manner.
Interestingly, we found their approach to be significantly more effective for larger seed size (Table~\ref{tb:2}). The APPNP~\cite{klicpera2018predict} model was a precursor to C\&S paper.
It consists of using a neural network to generate base predictions and then smoothing using power iterations of topic-sensitive page rank algorithm~\cite{haveliwala2003topic}, and the whole model is trained in an end-to-end manner.
It is similar to Pipeline1 of our model, which is also trained end-to-end, except that smoothing in Pipeline1 is done via a PDE.
They had no counterpart of Pipeline2. 
Another interesting but distinct hybrid model is from~\cite{wang2020unifying}, in which they propose to learn graph weights as a pre-processing step and then use those graph weights in a GCN for node classification.

\paragraph{NeuralODEs.} There has been some prior work of bringing NeuralODEs~\cite{chen2018neural} to do transductive learning on Graphs.
The earlier attempts using NeuralODE framework parameterized the derivative function with a few layers of GNN~\cite{poli2019graph,zhuang2019ordinary}.
The first attempt to do the transductive learning using a diffusion based method without parameterizing the derivative was in CGNN~\cite{xhonneux2020continuous}. 
\cmt{They proposed two models for independent feature channels and interacting feature channels using learnable weights.}
GRAND~\cite{chamberlain2021grand} went steps further and added the attention mechanism~\cite{vaswani2017attention} to learn the graph weights, graph rewiring after each backward pass, and proposed a non-linear version\cmt{ where the weights were changing during numerical integration}. 
It is worth noting that in all these models, the evolution of a PDE is done in a latent space. That is, the evolution is stacked between an encoder and decoder using MLPs.
Whereas, in our case, the evolution occurs in the actual space of the label vector $\mathbf{R^{k}}$ as required by~\eqref{eq:time-pde}; also, the Dirichlet boundary condition~\eqref{eq:bc} is respected during the evaluation on the validation and test sets.
Another important difference is that our model can incorporate new labels without retraining the model (Table~\ref{tb:2}).

\subsection{Conclusion}

\paragraph{Importance of parameter $\mathbf{{t}}$.} The parameter $t$ at which the evolution of~\eqref{eq:time-pde} is stopped is quite important. 
It acts like a scaling parameter of a variational problem. \cmt{\footnote{As t goes to infinity one gets the harmonic extension and for t equals zero one retains the learned front $\psi_0$.}} It was carefully tuned to yield the best accuracy and avoid reaching steady-state.
This is because at steady-state, one gets the harmonic extension, and the solution of the Dirichlet problem is unique\footnote{Assuming the graph is connected.} and independent of the initialization on the \mbox{unlabeled}~nodes. \cmt{but depends on the signal on the labeled nodes $g(u)$.}

\paragraph{Pros.} An essential advantage of our model is the ability to incorporate new labels without retraining the whole model (Table~\ref{tb:2}). It has a practical application for a huge graph where retraining would require considerable time.
It must be emphasized that GNNs and NeuralODEs do not have this advantage, although \cmt{Note that} this advantage was also a highlight of C\&S paper~\cite{huang2020combining}.
\ncmt{In terms of learnable parameters Row 08 (Table~\ref{tb:2}) is quite efficient. 
For example, on FMAsia dataset, Row 08 and GAT yield the same accuracy, but the latter requires 4 times as many parameters (more comparisons in the \mbox{Appendix~E}).
Row 09 does make the graph weights learnable, but this does not necessarily add large overhead in terms of time-taken (\mbox{Appendix~F}).}
Deeper GCNs are known to suffer from oversmoothing and cannot propagate information to longer range~\cite{li2018deeper}.
However, the proposed approach avoids this limiting behavior for $t$ tends to infinity.\footnote{Remark: In NeuralODE literature the \emph{Number Of Function Evaluations} is considered as a proxy for the depth.}
Finally, we believe the most significant advantage of this approach is simplicity, which makes it easier to debug.
For example, in the case of absurd predictions over some nodes, one could quickly check the learned front $\psi_0$ and interpret the absurdity in terms of initial probability.
This is not possible in the case of GNNs, or if the evolution of a PDE is stacked between encoder-decoder.

\paragraph{Societal impact.}  Trustworthiness in A.I. is a growing concern.
The key to using A.I. more responsibly to serve society is to use effective yet straightforward ML models. Table~\ref{tb:2} shows that our model is effective. It is simple as it basically corresponds to the PDE~\eqref{eq:time-pde}, and in this work, we show how to ameliorate its solution by learning the initial front on unlabeled nodes.

\paragraph{Future directions.}
\emph{Machine learning perspective}: 
for a very large and dense graph, learning graph weights directly may not be efficient.
One way to circumvent this issue is to bring the attention mechanisms~\cite{vaswani2017attention,velivckovic2017graph}  to learn similarities between the features.
\emph{PDE perspective}: the Dirichlet energy in~\eqref{eq:var-problem} corresponds to $p=2$ case of \emph{p}-Dirichlet energy, given as $ J_{p}(f) = \tfrac{1}{p} \sum_{u\in V}\|\nabla_{w}(f)(u)\|^{p} $.
It would be interesting to consider the variational problem~\eqref{eq:var-problem} for $p=1$ because the PDE corresponding to that would be anisotropic diffusion (edge-preserving), which yields piecewise constant results. Hence it is desirable as intuitively, the vector $f$ should be composed of zeros and ones.
Another interesting direction would be to consider the Dirichlet problem with novel Laplacians~\cite{calder2020properly,calder2018game,el2016asymptotic}, which effectively do semi-supervised learning for very small seed sizes. 

\section*{Acknowledgements} 
This work was done with the support of the French National Research Agency through the project \mbox{ANR-17-CE38-0004}.
The author thanks \mbox{Julien Rabin} for insightful discussions at the start of this work.

\bibliographystyle{named}
\small{
\bibliography{ijcai22}
}

\end{document}